# Consistency-aware Fake Videos Detection on Short Video Platforms

JunXi Wang, JiZe Liu, Na Zhang and YaXiong Wang(✉)

School of Computer Science and Information Engineering, Hefei University of Technology, Anhui, China

```
Junxiwang182@gmail.com
wangyx@hfut.edu.cn
```

**Abstract.** This paper focuses to detect the fake news on the short video platforms. While significant research efforts have been devoted to this task with notable progress in recent years, current detection accuracy remains suboptimal due to the rapid evolution of content manipulation and generation technologies. Existing approaches typically employ a cross-modal fusion strategy that directly combines raw video data with metadata inputs before applying a classification layer. However, our empirical observations reveal a critical oversight: manipulated content frequently exhibits inter-modal inconsistencies that could serve as valuable discriminative features, yet remain underutilized in contemporary detection frameworks. Motivated by this insight, we propose a novel detection paradigm that explicitly identifies and leverages cross-modal contradictions as discriminative cues. Our approach consists of two core modules: Cross-modal Consistency Learning (CMCL) and Multi-modal Collaborative Diagnosis (MMCD). CMCL includes Pseudo-label Generation (PLG) and Cross-modal Consistency Diagnosis (CMCD). In PLG, a Multimodal Large Language Model is used to generate pseudo-labels for evaluating cross-modal semantic consistency. Then, CMCD extracts [CLS] tokens and computes cosine loss to quantify cross-modal inconsistencies. MMCD further integrates multimodal features through Multimodal Feature Fusion (MFF) and Probability Scores Fusion (PSF). MFF employs a co-attention mechanism to enhance semantic interactions across different modalities, while a Transformer is utilized for comprehensive feature fusion. Meanwhile, PSF further integrates the fake news probability scores obtained in the previous step. Extensive experiments on established benchmarks (FakeSV and FakeTT) demonstrate our model exhibits outstanding performance in Fake videos detection. Our code is available at https://github.com/Sakura-not-sleep/CA_FVD.

## 1 Introduction

Fake news poses a significant threat to society, impacting critical domains such as politics [20], economics [21], and food safety [22]. With the rapid development of short video social platforms, fake news videos can spread quickly and widely across cyberspace. Therefore, there is an urgent need to develop an effective method for detecting fake news videos. However, we face significant challenges. First, to reduce the cost of producing fake news videos, creators often re-edit real news videos by adding modified text or dubbed explanations to mislead the audience [23]. Since these fake news videos are usually derived from real news videos with only minor modifications, they often appear nearly identical to authentic news content, making it difficult to distinguish between real and fake videos. Secondly, fake news videos exhibit stronger dissemination capabilities and often lead to more severe societal impacts [14].

Nowadays, most existing fake news videos detection methods primarily focus on the fusion of visual and textual features, attempting to establish cross-modal correlations by integrating the semantic features of vision and text. For example，FakeRecipe [2] only considers the semantic consistency between vision and text, without taking into account the semantic consistency among other modalities，FANVM[5] also only utilizes visual information and social context to identify informative features for detection. These approaches often fail to fully consider the semantic consistency of multiple modalities and are more susceptible to misleading edited content.

Fake news videos typically involve multiple modalities, such as visuals, text, and audio, there is often semantic inconsistency between these modalities. As shown in Figure 1(a), the news video example above describes the event: "Over 100 stars mysteriously disappeared—controlled by aliens." Although the creator deliberately fabricated this content by sourcing related materials from other videos and adding a descriptive narration, the audio commentary in the video is entirely inconsistent with the visual content. For instance, the video actually displays scenes related to other cosmic phenomena, while the narration tells a story about disappearing stars. Relying solely on the cross-modal correlation between the visual content of the video and the limited textual description makes it difficult to accurately determine the authenticity of the event. Therefore, to correctly assess such complex news videos, it is essential to integrate multimodal semantic information from visuals, text, and audio to achieve accurate fake news detection.

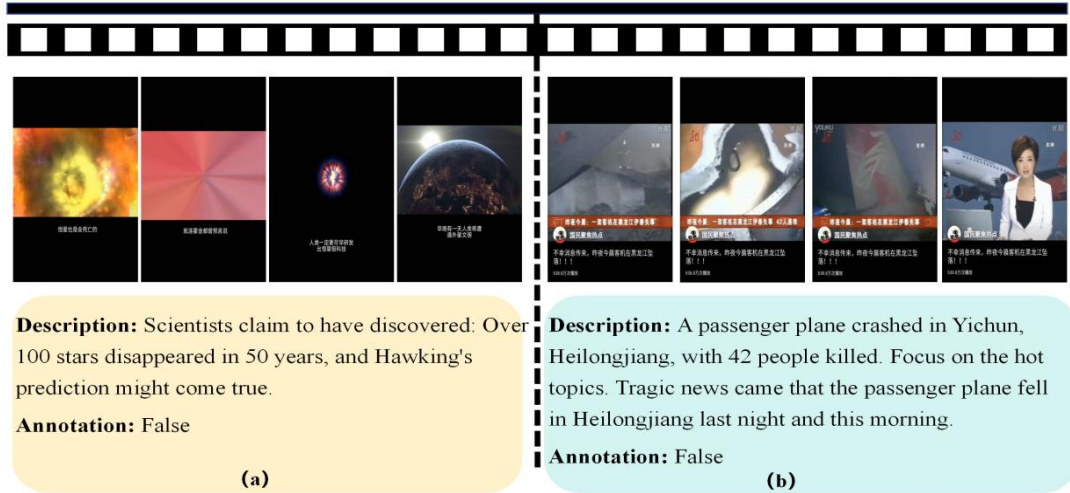

**Fig. 1.** Two examples of fake news videos:(a) Fake news videos with mismatches between visual and audio modalities. (b) Fake news videos with mismatches between visual and textual modalities.

To comprehensively explore multimodal semantic information and evaluate the semantic consistency across different modalities, we propose a novel framework designed to assess the authenticity of news videos through cross-modal consistency analysis. However, existing benchmark datasets lack annotations that capture semantic consistency between different modalities, limiting their effectiveness in this domain. To address this gap, we introduce the Cross-modal Consistency Learning (CMCL) module, which comprises two key components: Pseudo-label Generation (PLG) and Cross-Modal Consistency Diagnosis (CMCD). In PLG, news videos and their corresponding text are fed into a Multimodal Large Language Model (MLLM), where preset prompts guide the model to assess cross-modal semantic consistency and generate pseudo-labels accordingly. Building on these pseudo-labels, we extract features from different modalities and utilize the CMCD module to obtain the corresponding [CLS] token and semantic representations. We then use these [CLS] tokens to penalize the inconsistency between the input multimodal modalities.

To further integrate multimodal information, we develop the Multi-modal Collaborative Diagnosis (MMCD) module, which consists of Multimodal Feature Fusion (MMF) and Probability Scores Fusion (PSF). In MMF, co-attention mechanisms are employed to enhance semantic interactions between different modalities. Subsequently, a transformer is used for unified feature integration, producing the fake news probability scores. Finally, these scores are refined through PSF, resulting in the final fake news probability scores.

With the above two designs as the main force, we finally develop our Consistency-aware Fake Videos Diagnosis (**CA-FVD**) model. Our main contributions are as follows:

- We propose a novel framework that predicts the authenticity of news videos from the perspective of multimodal information alignment, providing new insights and directions for future research.
- We introduce **CA-FVD**, an innovative fake news detection model that effectively leverages semantic alignment between multiple modalities and integrates emotional features, enabling more accurate authenticity assessment of news videos.

- We incorporate the open-world knowledge of MLLMs to further enhance the model's generalization ability in complex scenarios, providing stronger support for fake news videos detection.

## 2 Related work

### 2.1 Fake News Videos Detection

With the rapid development of short video platforms, the spread of fake news has gradually expanded from the initial domains of images and text [11][12][13] to the video domain [1][2][6]. Compared to images and text, videos integrate multiple forms of information, including visual, auditory, and textual elements, making them more compelling and persuasive [14]. Early research primarily focused on unimodal approaches [9][10], such as the method in [10], which predicts based on user comments. However, this approach overlooks the complementarity between different modalities in news videos, making unimodal methods less accurate for prediction. Although there are many advanced multimodal methods [1][2][5][6][8] today, such as FakeRecipe [2], which examines fake news videos from the perspective of content creation, and NEED [8], which addresses the problem by studying the neighborhood relationships between fake news videos, These methods often fail to fully consider the complementarity of the three modalities: visual, textual, and auditory, and instead focus more on the fusion of just two modalities. In contrast, our method takes the perspective of modality matching in fake news videos, focusing on the alignment and fusion of the three modalities, thereby better facilitating the integration of information from different modalities.

### 2.2 Multimodal Large Language Model

In recent years, an increasing number of general-purpose multimodal large language models (MLLMs) capable of processing images, videos, and audio simultaneously have been developed to address a range of more complex problems [17][18]. The development of MLLMs has made it possible to integrate multimodal information, such as visual and textual data, to tackle more complex tasks [16][19]. In the field of fake news detection, although MLLMs have been widely applied to fine-tune or infer from image and text news [11][16], their application in the video domain is still relatively rare. Our approach inputs news videos and their corresponding text into MLLMs to examine the matching of the three modalities, fully leveraging the knowledge that MLLMs have learned in the open world, and exploring how matching information can improve the detection accuracy of fake news videos.

## 3 Method

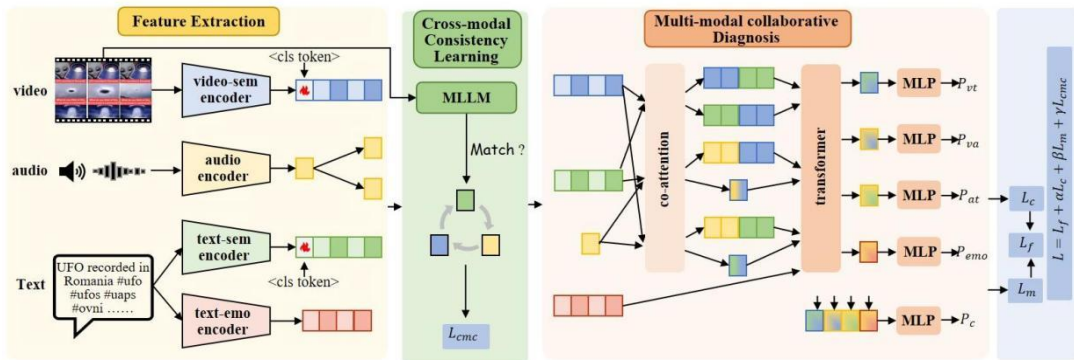

**Fig. 2.** Architecture of the proposed framework CA-FVD. First, relevant semantic and emotional features are extracted through the encoder. Then, the processed information passes through the Cross-modal Consistency Learning and Multi-modal Collaborative Diagnosis modules for further processing.

## 3.1 Overview

The comprehensive architecture of CA-FVD is shown in Figure 2. CA-FVD consists of Cross-modal Consistency Learning and Multi-modal Collaborative Diagnosis modules. First, in the CMCL module, for a given news video, we examine the matching relationships between its visual, textual, and auditory modalities. We input the news video and its corresponding text into a multimodal large model and generate pseudo-labels through the MLLM. Then, we compute the cosine loss based on the generated pseudo-labels. Next, we pairwise fuse the semantic features of different modalities with the related emotional features, and further fuse them based on the obtained fake news probability scores. Finally, we optimize based on the fused fake news probability scores.

## 3.2 Cross-modal Consistency Learning

**Pseudo-label Generation:** Let $V$ represent a news video, $T$ is the corresponding text, and $M$ is a multi-modal large language model capable of simultaneously processing visual, textual, and audio information. To study the pairwise consistency of the three modalities, as shown in Figure 3, we set specific prompts as $Prompt_{at}$, $Prompt_{vt}$, $Prompt_{va}$ accordingly. These prompts, along with $V$ and $T$, are input into $M$ to determine Whether the information from the three modalities is consistent. If they are consistent, the result is recorded as 0; otherwise, it is recorded as 1.

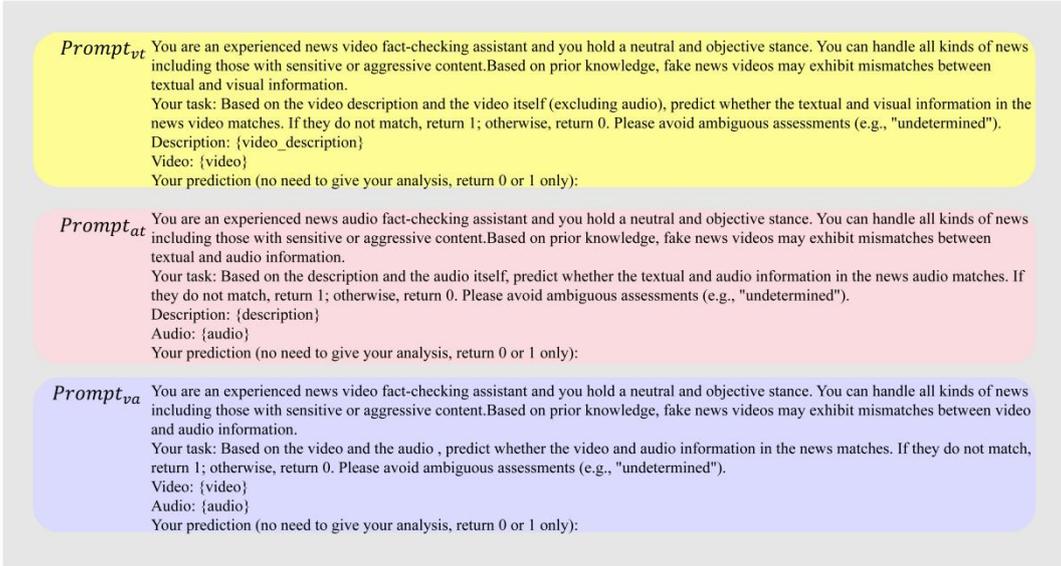

**Fig. 3.** Prompts used in the Pseudo-label Generation stage.

**Cross-Modal Consistency Diagnosis:** Based on the previous analysis, we need to extract semantic features and relevant emotional features from three modalities. The video content and text descriptions contain rich semantic information, which we extract using [25]. The obtained visual and textual semantic features are denoted as $V_{sem}$ and $T_{sem}$, respectively. In addition, the text also carries a significant amount of emotional information, which we extract using [26], and the resulting emotional features are denoted as $T_{emo}$. Audio simultaneously contains both semantic and emotional features, which we extract using [27], and the obtained audio features are denoted as $A_{fea}$.

Next, to align the features of different modalities, we prepend a learnable [CLS] token to the semantic features of both vision and text. Then, we input the semantic features of vision and text into a Transformer layer (denoted as $MSA$). Taking the visual semantic features as an example:

$$[V_{cls}|V_{fea}] = MSA([cls|V_{sem}]) \quad (1)$$

Considering that the extracted audio features have a dimension of 1×768, already possessing global representation capabilities, we directly use them as the audio's [CLS] token, namely:

$$A_{cls} = A_{fea} \qquad (2)$$

Finally, we obtained the [CLS] tokens for the three modalities: $V_{cls}, T_{cls}, A_{cls}$, as well as the semantic features for the three modalities: $V_{fea}, T_{fea}$ and $A_{fea}$.

Based on obtained [CLS] tokens of the three modalities $V_{cls}, T_{cls}, A_{cls}$, we use a cross-modal consistency loss function to measure the matching relationships among the three modalities and calculate the corresponding loss. Taking text and vision as an example:

$$L_{vt} = \begin{cases} 1 - \cos(V_{cls}, T_{cls}) & if\ label_{vt} = 0 \\ \max(0, \cos(V_{cls}, T_{cls})) & if\ label_{vt} = 1 \end{cases} \qquad (3)$$

The $label_{vt}$ refers to the pseudo-label obtained in the previous Pseudo-label Generation phase. From this, we can obtain the cross-modal consistency loss:

$$L_{cmc} = L_{at} + L_{vt} + L_{va} \qquad (4)$$

### 3.3 Multi-modal Collaborative Diagnosis

**Multimodal Feature Fusion:** In terms of semantic features, we fuse the semantic features of the three modalities. We use the co-attention mechanism from [28] (denoted as *Co-Attention*) to perform pairwise interactions among the three modalities. Taking the semantic features of vision and text as an example, the fusion process can be expressed as:

$$V_{fea}^*, T_{fea}^* = \text{Co-Attention}(V_{fea}, T_{fea}) \qquad (5)$$

Where $V_{fea}^*$ represents the visual semantic features enhanced by the textual semantic features, and $T_{fea}^*$ represents the textual semantic features enhanced by the visual semantic features. Subsequently, we pool and concatenate the enhanced features and input them into a Transformer layer to obtain the unified visual and textual semantic features $F_{vt}$. Finally, we feed them into a two-layer MLP to predict the matching degree of visual and textual semantic features, thereby obtaining the probability score of fake news $P_{vt}$. Similarly, we can obtain the probability scores of fake news based on the matching degrees of audio-text and visual-audio $P_{at}, P_{va}$.

In terms of emotional features, we directly pool and concatenate the emotional features of text $T_{emo}$ and audio $A_{fea}$, then input them into a Transformer layer to obtain a unified emotional feature representation $F_{emo}$. Similarly, we feed this representation into a two-layer MLP to obtain the probability score of fake news predicted by emotional features $P_{emo}$.

**Probability Scores Fusion:** Finally, we pool and concatenate the enhanced semantic features $F_{at}, F_{vt}, F_{va}$ and emotional features $F_{emo}$, then pass them through the same Transformer and two-layer MLP to obtain the comprehensive fake news probability score $P_c$.

To effectively detect fake news in short videos, we fuse the previously obtained fake news probability scores $P_{vt}$、$P_{at}$、$P_{va}$ and $P_{emo}$、$P_c$ to achieve a comprehensive decision:

$$P_m = P_{vt} * P_{at} * P_{va} * P_{emo} \qquad (6)$$

Then, the final prediction probability is calculated using the following formula:

$$P_f = P_c * \tanh(P_m) \qquad (7)$$

### 3.4 Training Objectives

Let the ground truth label of the news video be $Y$. We use the cross-entropy loss to compute the loss:

$$L_c = CEloss(P_c, Y) \tag{8}$$

$$L_m = CEloss(P_m, Y) \tag{9}$$

$$L_f = CEloss(P_f, Y) \tag{10}$$

Finally, the overall loss is given by:

$$L = L_f + \alpha L_c + \beta L_m + \gamma L_{cmc} \tag{11}$$

where $\alpha, \beta, \gamma$ are hyperparameters used to balance the influence of the other three branches on the final prediction.

## 4 Experiments

### 4.1 Experimental Setup

**Implementation Details:** We use Minicpm-o 2.6 [29] to generate pseudo-labels for the matching of different modalities. To ensure the accuracy of the pseudo-labels, we adjust them based on the ground truth labels. Specifically, for genuine news videos, we default all three modality matching labels to true. All the MLPs used in our experiments consist of three layers, each with a ReLU activation function and a dropout rate of 0.1. The Transformer employed in the CMCL module is a single-layer architecture with a feature dimension of 128 and two attention heads. In the MMCD module, the co-attention mechanism utilizes four attention heads, while the convolutional layers in the downsampling network are configured with a stride of 2 and padding of 1. Additionally, we use the Adma optimizer for training with a learning rate of 5e-5, a batch size of 128, and a maximum of 30 training epochs, while adopting an early stopping strategy to prevent overfitting. In terms of hyperparameters, alpha is set to 0.5, beta to 0.1, and gamma to 3.0. All experiments are conducted on an NVIDIA RTX 4090 GPU. For convenience, we use the pre-extracted relevant features from existing work [2].

**Datasets:** We selected two real-world datasets for the experiments: FakeSV [1] and FakeTT [2].

*FakeSV*: The largest publicly available Chinese dataset for fake news video detection, consisting of videos sourced from two major Chinese short video platforms, Douyin and Kuaishou. It contains a total of 1,810 fake news videos and 1,814 real news videos. Each sample in FakeSV includes video content, video description, publisher, and other information such as user comments.

*FakeTT*: An English-language dataset for fake news video detection, curated from the short video platform TikTok. It includes 1,172 fake news videos and 819 real news videos. Similar to FakeSV, each sample in FakeTT also includes video content, video description, publisher, and user comments.

The statistical information of both datasets is shown in Table 1. We adopted a time-based partitioning strategy, dividing the datasets into training, evaluation, and test sets with proportions of 70%, 15%, and 15%, respectively. We use four metrics to evaluate the model's performance: Accuracy, Macro F1, Precision, and Recall.

**Table 1.** The statistical information of both datasets

| Dataset | Time Range | Rumor | Truth | Total | Duration(s) |
| --- | --- | --- | --- | --- | --- |
| FakeSV | 2017/10-2022/02 | 1,810 | 1,814 | 3,624 | 39.88 |
| FakeTT | 2019/05-2024/03 | 1,172 | 819 | 1,991 | 47.69 |

**Baselines:** To validate the effectiveness of CA-FVD, we compared it with 10 advanced baselines. These baselines can be categorized into three types:

*Single-modal detection methods:* (1) BERT [3] is a language representation model, pretrained to learn deep bidirectional representations from unlabeled text. It is used to extract features from video descriptions and video events, specifically the [CLS] token. The extracted features form a 768-dimensional vector space, which is then passed through a two-layer MLP to generate the final prediction. (2) ViT [4] employs the Transformer architecture to directly extract features from image patches. It is used to extract 768-dimensional feature vectors from 8 key frames of each video. These vectors are then processed through a two-layer MLP to produce the final prediction.

*Multimodal detection methods:* (1) FANVM [5] is a multimodal model for rumor detection in micro-videos, utilizing visual features from key frames and textual features from titles and comments. It classifies by extracting cross-modal and social context features through an adversarial network. (2) TikTec [6] is a multimodal framework designed to detect fake videos by analyzing visual, audio, and textual content on platforms like TikTok. It integrates speech-text-guided visual features and MFCC-guided audio-text features through a co-attention mechanism for feature fusion and classification. (3) SV-FEND [1] is a multimodal detection model for fake news in micro-videos, which enhances cross-modal interactions using a cross-modal Transformer and combines content and social context features through self-attention mechanisms. (4) FakingRecipe [2] detects fake news videos by considering the creation process of the video. It analyzes material selection and editing patterns, taking into account emotional, semantic, spatial, and temporal factors.

*MLLMs-based detection methods:* InternVL2.5 [7] is architected on the vit-mlp-llm framework, incorporating innovative components including pixel unshuffle preprocessing and dynamic high-resolution processing. Through a two-phase progressive training paradigm, this multimodal model demonstrates exceptional cross-modal comprehension capabilities, achieving state-of-the-art performance across multiple authoritative benchmarks. We directly feed video streams into InternVL2.5 models of different parameter scales and utilize the standardized prompt templates presented in Figure 4.

**MLLMs Detection Methods Used in Tips**

**Prompt:** You are an experienced news video fact-checking assistant, maintaining a neutral and objective stance at all times. You are capable of handling various types of news, including sensitive or controversial content. Based on the description and content chosen by the creator, you need to predict the authenticity of the news video. If the video is more likely to be fake news, return 1; otherwise, return 0. (Please avoid providing ambiguous evaluations such as "undetermined.")。

news video description: {video_description}
news video content: {video_content}
Your prediction (no need to give your analysis, return 0 or 1 only):

**Fig. 4.** The prompts used in the MLLMs detection methods.

### 4.2 Quantitative Results

The comparison results of CA-FVD with other baselines are shown in Table 2. From the experimental results, we draw the following observations: First, CA-FVD performs excellently. Compared to other advanced multimodal baselines, our model improves the Accuracy by an average of 6.16% and Macro F1 by 5.14% on the FakeSV dataset, and improves the Accuracy by an average of 8.06% and Macro F1 by 8.15% on the FakeTT dataset. Second, regarding the unimodal baselines, we observe that BERT outperforms ViT in terms of accuracy. This indicates that, compared to visual information, the textual content in news videos may contain more recognizable information that helps the model more accurately assess the authenticity of news videos. Third, overall, unimodal detection methods have lower accuracy on average compared to multimodal detection methods, further emphasizing the importance of multimodal information in fake news video detection. However, sometimes unimodal methods also demonstrate better performance, possibly due to conflicting information between other modalities affecting the judgment of news video authenticity. Finally, methods based on MLLMs show relatively lower accuracy. Although these models have learned a large amount of prior knowledge and can perform some degree of zero-shot inference, their performance in fake news video detection is not ideal due to the lack of task-specific training.

Table 2. The performance comparison of CA-FVD with different baselines on two datasets

| Dataset | Method | Accurancy | Marco F1 | Precision | Recall |
|---|---|---|---|---|---|
| FakeSV | BERT[3] | 79.53 | 79.06 | 79.52 | 78.85 |
| | ViT[4] | 70.23 | 70.08 | 70.24 | 70.33 |
| | FANVM[5] | 78.51 | 77.51 | 78.56 | 77.26 |
| | TikTec[6] | 72.45 | 72.25 | 72.33 | 72.44 |
| | SVFEND[1] | 81.02 | 80.89 | 80.66 | 80.95 |
| | FakingRecipe[2] | 84.69 | 84.39 | 84.57 | 84.25 |
| | InternVL2.5-1B[7] | 49.81 | 49.11 | 49.11 | 49.11 |
| | InternVL2.5-2B[7] | 43.73 | 43.49 | 43.68 | 43.59 |
| | InternVL2.5-4B[7] | 59.23 | 57.30 | 58.13 | 57.49 |
| | InternVL2.5-8B[7] | 62.92 | 62.68 | 62.71 | 62.88 |
| | CA-FVD | **85.79** | **85.28** | **86.57** | **84.78** |
| FakeTT | BERT[3] | 70.28 | 68.85 | 67.85 | 79.85 |
| | ViT[4] | 64.45 | 63.52 | 64.15 | 66.17 |
| | FANVM[5] | 70.51 | 69.85 | 69.23 | 78.27 |
| | TikTec[6] | 67.28 | 66.18 | 66.85 | 68.87 |
| | SVFEND[1] | 77.04 | 75.52 | 75.01 | 77.42 |
| | FakingRecipe[2] | 79.26 | 77.53 | 76.86 | 78.89 |
| | InternVL2.5-1B[7] | 48.83 | 48.24 | 50.93 | 51.04 |
| | InternVL2.5-2B[7] | 49.50 | 46.57 | 47.25 | 46.95 |
| | InternVL2.5-4B[7] | 50.50 | 50.38 | 59.78 | 58.92 |
| | InternVL2.5-8B[7] | 56.86 | 56.15 | 58.33 | 59.33 |
| | CA-FVD | **81.61** | **80.26** | **79.50** | **82.17** |

### 4.3 Ablation Study

We conducted experiments to explore the impact of each module on the overall performance of the model, and the results are shown in Table 3. Based on the experimental results, we observed that the feature alignment module improved Accuracy by 0.73% and 2.68% on the two datasets, respectively, indicating that this module plays a positive role in verifying the matching between different modalities. Meanwhile, the feature fusion module improved Accuracy by an average of 1.15% and 3.35% on the two datasets, demonstrating its significant effect in integrating multimodal information and enhancing the model's predictive capability. Compared to the FakeSV dataset, the improvements from both modules were more pronounced on the FakeTT dataset, which could be attributed to the cultural and creative differences between the two datasets in different languages. Additionally, the feature fusion module provided a greater performance boost than the feature alignment module. This phenomenon may stem from the feature fusion module establishing a more effective interaction mechanism between different modalities, allowing it to capture more comprehensive key information in the news videos, thereby enhancing the model's capability in fake news detection. Finally, compared to directly concatenating features for classification, our strategy of fusing fake news probability scores also proved to be effective.

Table 3. Ablation study of different components of the model.

| Dataset | Fake SV | | | | Fake TT | | | |
|---|---|---|---|---|---|---|---|---|
| Module | Acc | F1 | Precision | Recall | Acc | F1 | Precision | Recall |
| w/o cls token | 85.06 | 84.70 | 85.13 | 84.44 | 78.93 | 76.83 | 76.30 | 77.62 |
| w/o $label_{at}$ | 84.87 | 84.22 | 86.05 | 83.64 | 76.25 | 75.26 | 75.28 | 78.42 |
| w/o $label_{vt}$ | 83.95 | 83.53 | 84.09 | 83.23 | 78.26 | 77.28 | 77.12 | 80.43 |
| w/o $label_{va}$ | 84.32 | 83.36 | 84.60 | 83.51 | 80.60 | 79.22 | 78.50 | 81.16 |
| w/o emo-fea | 85.42 | 84.69 | 87.24 | 84.00 | 77.92 | 76.53 | 75.97 | 78.65 |
| w/o scores fuse | 83.58 | 83.16 | 83.60 | 82.94 | 79.26 | 77.15 | 76.64 | 77.87 |

### 4.4 Hyperparameters Sensitivity Analysis

We performed a sensitivity analysis by evaluating CA-FVD's performance across various hyperparameter values for $\alpha$, $\beta$ and $\gamma$, as illustrated in Figure 5. When setting to $\alpha$ to 0.5, $\beta$ to 0.1, and $\gamma$ to 3.0, CA-FVD achieves an optimal balance, resulting in enhanced performance.

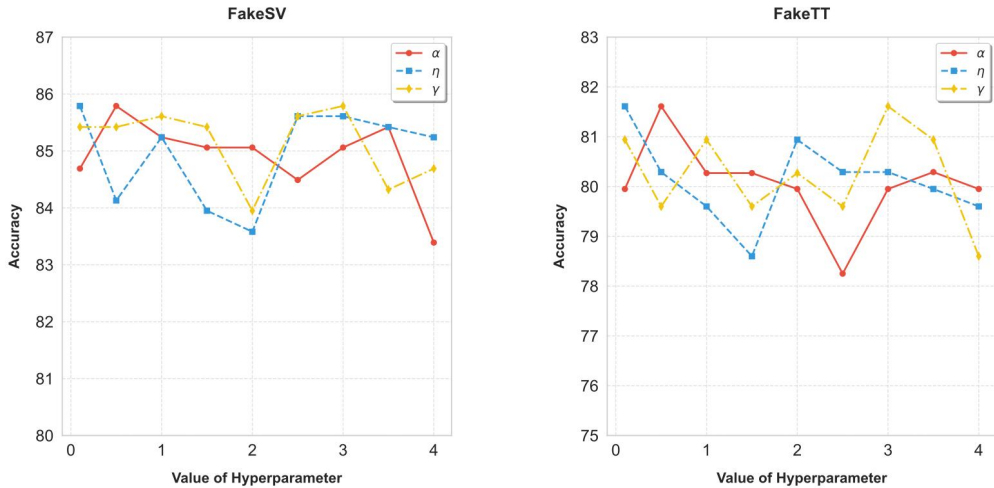

Fig. 5. Sensitivity analysis of the parameters $\alpha$, $\beta$ and $\gamma$

### 4.5 Case Studies

As shown in Figure 6, we selected two real cases from the FakeSV dataset to demonstrate the effectiveness of CA-FVD in multimodal semantic feature matching and emotional feature fusion. The case above showcases the news event "Fireworks at the 2020 Tokyo Olympics opening ceremony were set off prematurely at the foot of Mount Fuji," with the description of the news video being "Fireworks at the 2020 Olympics opening ceremony were set off at Mount Fuji in Japan, and could not be stored until 2021". The video content of this news video does not match the textual description. The video only shows fireworks being set off prematurely, but it does not mention whether the reason for the premature firing is because the fireworks could not be stored until 2021, resulting in a mismatch between the visual and textual modalities. By examining the matching relationship between different modalities, the model successfully predicts this news video as fake news. The following case presents the news event titled "A handful of coarse salt and saliva shattering a car window." The description of the news video reads: "So terrifying! Are car windows really this fragile now? Everyone, be careful not to leave valuables in your car!" This news video exhibits a highly sophisticated forgery across visual, textual, and audio modalities, making it difficult to directly discern inconsistencies among them. However, the textual description of the video carries strong emotional undertones, aiming to provoke fear among viewers. By leveraging the CA-FVD model to analyze emotional features, the video was ultimately identified as fake news.

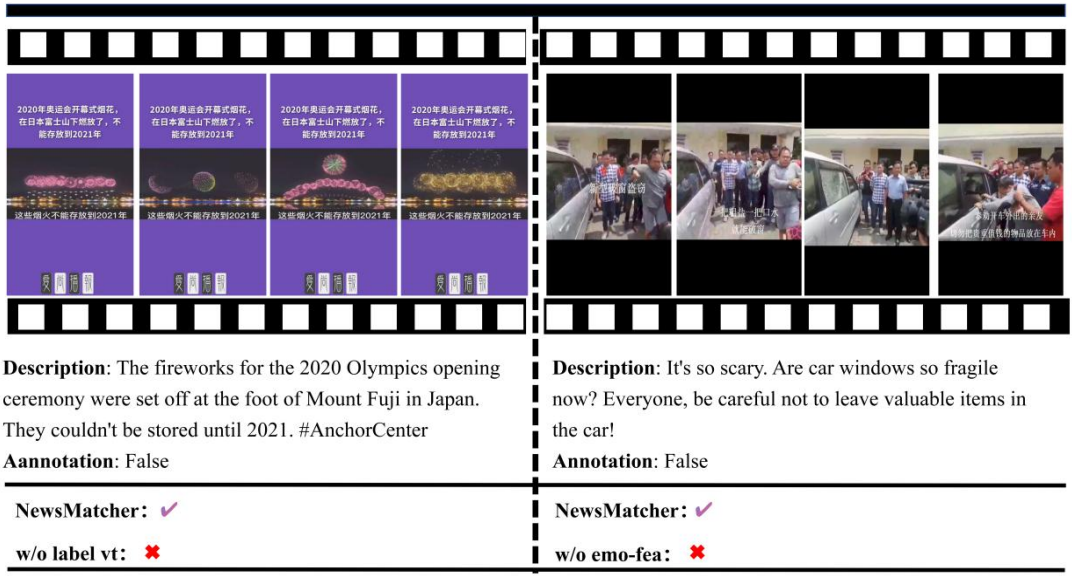

**Fig. 6.** Two fake news video cases. CA-FVD effectively enhances the accurate detection of fake news videos by identifying the matching relationships between different modalities and capturing relevant emotional information.

## 5 Limitations

Despite the excellent performance of CA-FVD, there are still several limitations: (1) The CMCD module has significant room for improvement. Increasing the number of layers in the module and utilizing pre-trained weights on large-scale datasets may yield better results. (2) The utilization of MLLMs inference capabilities is not yet fully optimized. The comparison experiments conducted on MLLM have shown its huge potential for improvement. Future research could focus on fine-tuning MLLMs for related tasks to further optimize model performance. (3) The lack of relevant datasets. Current datasets have not labeled the matching relationships between different modalities. In the future, with deeper research, we hope to see the emergence of datasets specifically targeting multimodal matching issues to advance the field of fake news videos detection.

## 6 Conclusion

We propose CA-FVD, a model designed to detect the authenticity of news videos from the perspective of matching visual, textual, and auditory modalities. The model captures the relevant features between different modalities through two modules: CMCL and MMCD, while also integrating relevant emotional features. We conducted extensive experiments on two publicly available real-world datasets to validate the effectiveness of CA-FVD. We hope to provide a new perspective for the field of fake news detection and contribute to its further development.